\documentclass[11pt]{article}

\usepackage[margin=1in]{geometry}
\usepackage{graphicx}
\usepackage{pdflscape}
\usepackage{xcolor}
\usepackage{amsmath,amssymb}
\usepackage{graphicx}
\usepackage[T1]{fontenc}
\usepackage[utf8]{inputenc}
\usepackage{lmodern}
\usepackage{microtype}
\usepackage{booktabs}
\usepackage{longtable}
\usepackage{array}

\usepackage{setspace}
\usepackage[numbers,sort&compress]{natbib}
\usepackage{hyperref}
\hypersetup{
  pdftitle={VibrantSR: Sub-Meter Canopy Height Models from Sentinel-2 Using Generative Flow Matching},
  pdfauthor={Vibrant Planet},
  hidelinks,
  colorlinks=true,
  linkcolor=blue,
  citecolor=blue,
  urlcolor=blue
}

\usepackage{placeins} 

\usepackage{fancyhdr}
\fancypagestyle{firstpage}{
  \fancyhf{}
  \fancyfoot[L]{\footnotesize\rule{0.35\textwidth}{0.4pt}\\[0.3em]
  $^{*}$Corresponding author: \textbf{kiarie@vibrantplanet.net}\\
  $^{\dagger}$Work carried out while at Vibrant Planet.}

}

\title{VibrantSR: Sub-Meter Canopy Height Models from Sentinel-2 Using Generative Flow Matching}

\author{
Kiarie Ndegwa$^{*}$, Andreas Gros$^{\dagger}$, Tony Chang, David Diaz, Vincent A. Landau, \\
Nathan E. Rutenbeck, Luke J. Zachmann, \\
Guy Bayes$^{\dagger}$, Scott Conway \\[1em]
Vibrant Planet Public Benefit Corporation, Truckee, CA, USA
}

\date{January 2026}

\begin{document}
\maketitle
\thispagestyle{firstpage}

\begin{abstract}
We present VibrantSR (Vibrant \textbf{S}uper-\textbf{R}esolution), a generative super-resolution framework for
estimating 0.5~meter canopy height models (CHMs) from 10~meter
Sentinel-2 imagery. Unlike approaches based on aerial imagery that are constrained by infrequent
and irregular acquisition schedules, VibrantSR leverages globally available
Sentinel-2 seasonal composites, enabling consistent monitoring at a seasonal-to-annual cadence. 
Evaluated across 22 EPA Level 3 eco-regions in the
western United States using spatially disjoint validation splits,
VibrantSR achieves a Mean Absolute Error of 4.39~meters for canopy
heights $\ge$2~m, outperforming Meta (4.83~m), LANDFIRE (5.96~m), and
ETH (7.05~m) satellite-based benchmarks. While aerial-based VibrantVS
(2.71~m MAE) retains an accuracy advantage, VibrantSR enables
operational forest monitoring and carbon accounting at continental
scales without reliance on costly and temporally infrequent aerial acquisitions.

\noindent\textbf{Keywords:} canopy height model, canopy height super-resolution, flow matching,
Sentinel-2, lidar, remote sensing, forest structure
\end{abstract}

\section{Introduction}\label{sec:intro}

High-resolution canopy height models (CHMs) are a foundational
geospatial layer for characterizing forest structure and its change over
time. Sub-meter CHMs support operational tasks that depend on fine
spatial detail---such as canopy gaps, edges, and within-stand
heterogeneity---including fuels and wildfire planning \cite{north2021pyrosilviculture,cruz2005development}, biomass and carbon accounting \cite{coops2021modelling}, habitat assessment \cite{vierling2008lidar}, and post-disturbance
monitoring \cite{stevens2018evidence}. Many of these applications
additionally require repeatable updates at a seasonal-to-annual cadence
across large regions, rather than one-off, static maps \cite{wulder2008role}.

Airborne lidar remains the most direct and accurate source of canopy
height mapping.
However, lidar acquisitions are spatially
fragmented, costly to repeat frequently, and often temporally misaligned
with rapidly evolving forest conditions driven by disturbance,
management, and climate variability \cite{bolton2020optimizing}. This creates a
persistent trade-off between high-fidelity structural measurements where
lidar exists and the need for broad, temporally consistent coverage that
aligns with real-world monitoring and decision-making cycles \cite{hummel2011comparison}. Freely available satellite missions such as Sentinel-2
partially address the latter requirement by providing globally
consistent multispectral imagery with regular revisit, but their native
10--60~m spatial resolution fundamentally under-resolves many canopy
structural patterns that are critical at management scales \cite{drusch2012sentinel}.

Recent machine learned canopy-height products illustrate this tension between accessible data and high spatial resolution requirements \cite{tolan2024very,landfire2022,lang2023high}.
Global and continental-scale models trained using combinations of spaceborne and
airborne lidar labels offer broad coverage, but typically operate at
meter to tens-of-meters output resolution and rely on regression
objectives that smooth out fine-scale canopy structure. While these
approaches are effective for large-area summaries, the resulting
predictions often suppress canopy gaps and attenuate height variability,
distorting the structural distributions that downstream ecological and
fire-behavior models depend on. As a result, while aerial imagery-based approaches set an upper bound on achievable
accuracy for modeled canopy height, they do not fully satisfy the operational requirement for
consistent, repeatable annual monitoring.

In our prior work, VibrantVS, we showed that high-resolution NAIP (National Agriculture Imagery Program) aerial
imagery can support accurate 0.5~m canopy height estimation across
diverse western United States eco-regions \cite{chang2025vibrantvs}.
However, dependence on NAIP aerial imagery introduces practical
constraints that include irregular and regionally
heterogeneous acquisition schedules, infrequent revisit cadence (usually 2 years), and radiometric and geometric inconsistencies driven by flight-path
variation and look-angle effects even within a single year. These factors limit suitability for
frequent refresh cycles and global-scale deployment.

We introduce \textbf{VibrantSR} (\emph{Vibrant Super-Resolution}), a
generative framework for producing 0.5~m CHM from 10~m Sentinel-2 imagery. VibrantSR first derives compact
representations of the Sentinel-2 inputs using separately
pretrained, then frozen, feature extractors. A generative mapping then
translates these Sentinel-2 representations into a high-resolution CHM
representation learned from 0.5~m lidar-derived training targets, which
is subsequently decoded into a 0.5~m CHM. By posing canopy-height
inference as learning the distribution of plausible fine-scale structure
conditioned on Sentinel-2, rather than predicting a single deterministic
surface via pixel-wise regression, VibrantSR is encouraged to preserve
realistic canopy variability that is often smoothed by existing
satellite-based approaches. Full model and training details are provided
in Section~\ref{sec:method}.

We evaluate VibrantSR on the western United States using spatially
disjoint train and validation splits across EPA Level~3 eco-regions
\cite{epa2013level}. We focus on this region to ensure continuity with
our prior VibrantVS benchmark, which was constructed exclusively over
the western United States, enabling direct comparison under a consistent data,
preprocessing, and evaluation protocol. We benchmark against VibrantVS as well as widely used
satellite imagery-based CHM products \cite{tolan2024very,landfire2022,lang2023high}.

Our objective is to demonstrate whether VibrantSR can offer reduced error
compared to existing satellite-derived products for canopy heights $\ge$2~m
while producing sharper, higher fidelity representations of
canopy structure at sub-meter resolution and benchmark performance when compared to other well known modeled CHM products. 

\section{Related Work}\label{sec:related}

\subsection{Learning-Based Canopy Height Estimation}\label{sec:related-learning}

Deep learning methods for CHM estimation have evolved from convolutional
neural networks to vision transformers \cite{tolan2024very,lang2023high,wagner2024sub}. Early approaches used regression losses to
predict pixel-wise height values from aerial or satellite imagery. More
recent work has explored self-supervised pretraining \cite{tolan2024very} and multi-task learning \cite{chang2025vibrantvs} to improve
generalization across diverse forest types. Wagner et al.~\cite{wagner2024sub} demonstrated sub-meter tree height
mapping using aerial images with a U-Net architecture informed by lidar.
Our previous work, VibrantVS \cite{chang2025vibrantvs}, introduced a
multi-task vision transformer trained on NAIP aerial imagery that
achieved state-of-the-art accuracy for CHM estimation in the western
United States.

\subsection{Operational CHM Products}\label{sec:related-products}

Several operational CHM products are available for large-scale forest
monitoring. These models adopt a regression-based canopy height estimation approach and typically predict conditional mean height surfaces from multispectral and ancillary inputs. Meta \cite{tolan2024very} uses Maxar Vivid2 imagery (0.5~m
resolution) with a DINOv2-based architecture trained on NEON (National Ecological Observatory Network) lidar
\cite{thibault2023us} and GEDI (Global Ecosystem Dynamics Investigation) data, providing 1 m resolution
global coverage for 2020. LANDFIRE \cite{landfire2022} employs a
regression-tree based method using Landsat multi-spectral data along with
topography and biophysical features at 30~m resolution with
periodic updates. ETH \cite{lang2023high} provides a deep learning
ensemble trained on Sentinel-2 imagery with sparse GEDI lidar labels at 10~m resolution with global coverage.

\subsection{Super-Resolution and Generative Models for Remote Sensing}\label{sec:related-generative}

Recent work has demonstrated that sub-meter predictions can be derived
from Sentinel-2 imagery despite its 10~m native resolution. Sirko et al.~\cite{sirko2023high} achieved 0.5~m resolution building and
road segmentation from Sentinel-2 using teacher-student distillation,
attaining 78.3\% mIoU for building detection compared to 85.3\% from
high-resolution imagery. This work established that fine-scale spatial
structure can be recovered from coarse satellite inputs when sufficient
training signal is available.

More recently, generative models based on diffusion and flow-based formulations have been adopted for remote sensing super-resolution and image synthesis. These efforts have improved training stability and the ability to model complex, multi-modal output distributions compared to Generative Adversarial Networks (GANs). Latent diffusion models have been applied to Sentinel-2 super-resolution and harmonization tasks, where operating in a compressed latent space substantially reduces computational cost while preserving spatial and spectral fidelity ~\cite{aybar2024opensr, ebel2020sen12mscr, lanaras2018super}. 

In parallel, diffusion and score-based models have been explored for the generative downscaling of continuous geophysical fields in climate and weather applications, where the objective is not only visual realism but the preservation of physically meaningful spatial variability and distributional properties ~\cite{harris2022generative, price2022generative}. Diffusion models and flow matching approaches offer advantages over GANs in training stability and sample diversity ~\cite{liu2023flow,lipman2023flow}. Recent work has shown that rectified flow models scale predictably to high-resolution image synthesis when paired with a transformer backbone ~\cite{esser2024scaling}, demonstrating their suitability for high-capacity generative tasks. However, existing applications of flow matching and diffusion in Earth observation have largely focused on image-domain outputs or same-modality super-resolution, rather than the generation of biophysical structure fields.

To our knowledge there has not been any work adopting a latent generative formulation to model a full conditional distribution of fine-scale canopy structure based on a multispectral satellite image, thus motivating our adoption of this framework for CHM super-resolution.

\section{Study Area and Data}\label{sec:data}
This study builds directly on the study area definition and spatial data partitioning introduced in our prior work, VibrantVS \cite{chang2025vibrantvs}. We therefore summarize only the key aspects relevant to VibrantSR and refer readers to Chang et al.~\cite{chang2025vibrantvs} for full details.

\subsection{Sentinel-2 Imagery}\label{sec:data-s2}
VibrantSR uses 12-band Sentinel-2 Level-2A surface reflectance imagery
\cite{drusch2012sentinel}, specifically the Sentinel-2-C1-L2A collection
distributed by Element84 \cite{element84_sentinel2_c1_l2a}. We aggregated
observations via a temporal median over the 2024 temperate summer months
(June--August) after masking individual acquisitions to remove clouds using the Scene Classification Layer (SCL) that is served alongside the data.

This temporal median aggregation reduces transient artifacts (e.g., residual
cloud contamination) and captures vegetation structure during the peak temperate growing season. Inputs from all Sentinel-2 bands were processed and resampled to 10 m resolution using nearest neighbor interpolation;
each training sample consisted of a $48\times48$ Sentinel-2 image
($480\,\mathrm{m}\times480\,\mathrm{m}$) at 10 m resolution paired with a corresponding
$960\times960$ CHM tile at 0.5 m resolution.
Sentinel-2 inputs were then standardized to zero mean and
unit variance based on dataset statistics. 

\subsection{Ground-Truth CHMs}\label{sec:data-chm}

Ground-truth CHMs were derived from United States USGS 3D Elevation
Program (3DEP) airborne lidar acquisitions \cite{usgs20243dep}. CHMs were computed as the difference between the digital surface
model (DSM) and digital terrain model (DTM), representing height above
ground \cite{vanleeuwen2010retrieval}. All metrics are provided for
canopy heights at or above 2~m, following common practice in lidar applications in forest ecosystems to differentiate tree canopy from shrub and other non-tree vegetative cover. The preprocessing of these groud-truth CHMS were
minimal: we applied min-max normalization to rescale height values to the
target range of the autoencoder. We excluded tiles containing negative height values or open water to restrict training and evaluation to terrestrial vegetation. Heights exceeding biologically plausible limits were clipped to a maximum of 120 m to ensure stable training.

\subsection{Evaluation Protocol}\label{sec:data-eval}

\begin{figure}[t]
\centering
\includegraphics[width=\textwidth]{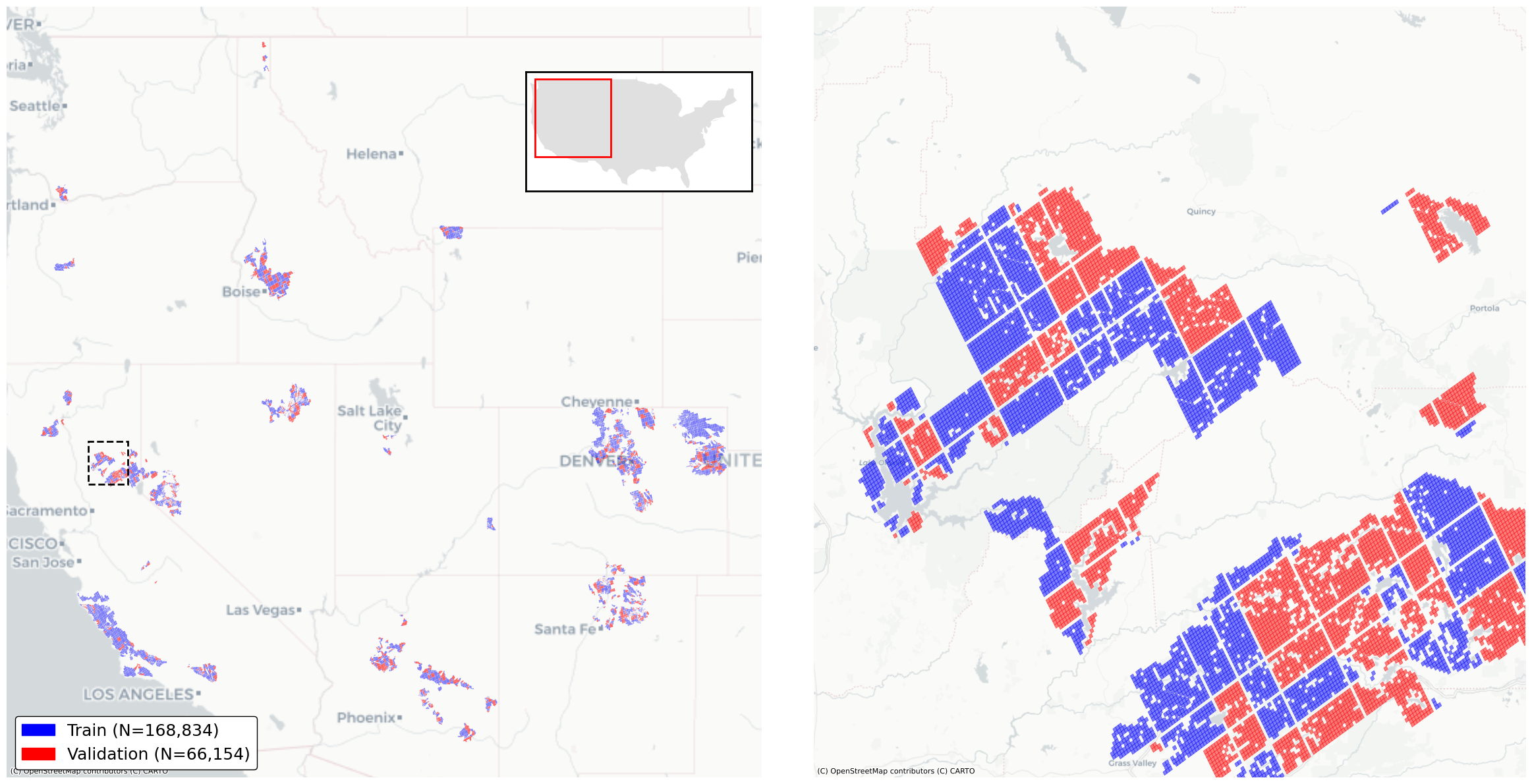}
\caption{Spatial partitioning of training and validation tiles across the western United States. The left panel shows the distribution of tiles assigned to the training (blue) and validation (red) folds. The right panel provides a regional zoom illustrating strict spatial separation between folds.}
\label{fig:folds}
\end{figure}

\begin{figure}[t]
\centering
\includegraphics[width=\textwidth]{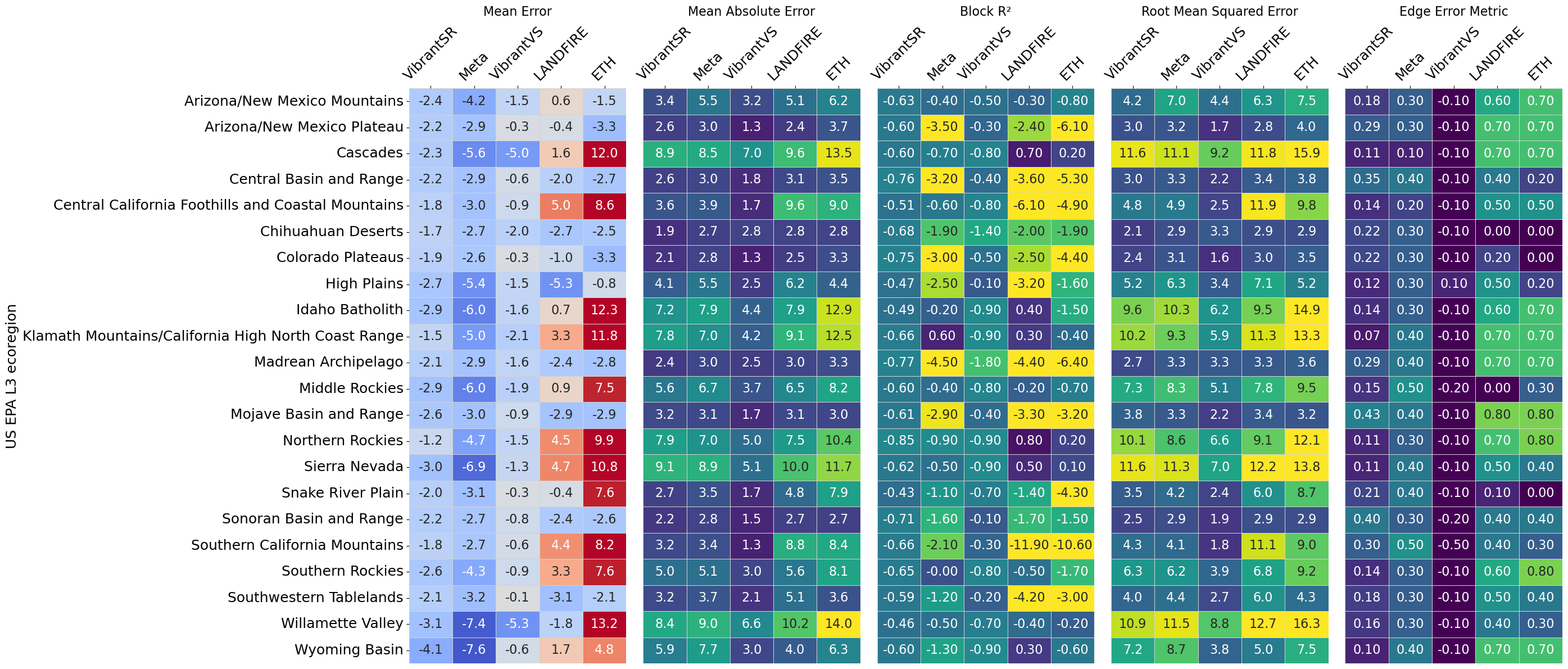}
\caption{Error metrics across EPA Level~3 ecoregions.}
\label{fig:eco-regions}
\end{figure}

We employed a spatially disjoint ``checkerboard'' split to ensure the
model was evaluated on geographically distinct areas (Fig.~\ref{fig:folds}). 
This prevents spatial autocorrelation from inflating performance metrics. 
The final dataset was comprised of 168,834 training tiles and 66,154 validation tiles over 4,135 square kilometers, 
spanning 22 EPA Level 3 eco-regions \cite{epa2013level} 
in the western United States and ensuring evaluation across diverse forest types,
climate zones, and topographic conditions (Fig.~\ref{fig:eco-regions}).





\section{Methods}\label{sec:method}
\subsection{Modeling Description}\label{sec:modeling-description}
\begin{figure}[htbp]
\centering
\includegraphics[width=\textwidth]{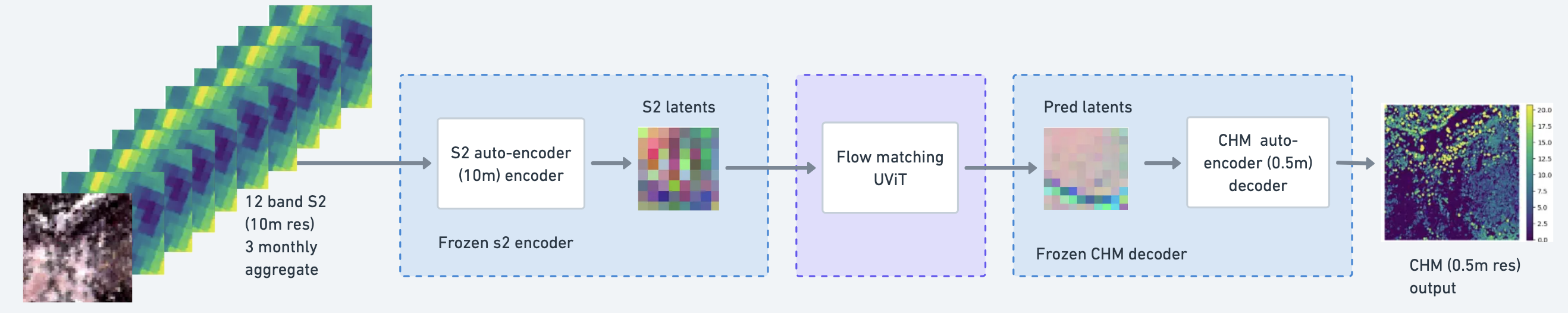}
\caption{VibrantSR architecture showing the flow from Sentinel-2 inputs through frozen autoencoders and the trainable flow matching network to CHM outputs.}
\label{fig:arch}
\end{figure}

VibrantSR employs a generative flow matching architecture operating in
compressed latent space (Fig.~\ref{fig:arch}). The pipeline
consists of three components: (1) a frozen Sentinel-2 autoencoder that
compresses 12-channel inputs into a compact latent representation; (2) a
trainable flow matching network that transforms Sentinel-2 latents into
CHM latents; and (3) a frozen CHM autoencoder decoder that reconstructs
0.5-meter predictions from the generated latents.

The flow matching network is a U-shaped Vision Transformer (U-ViT)
configured for velocity prediction. The architecture uses a linear flow
path and was trained to minimize the velocity matching loss between the
source (Sentinel-2 latent) and target (CHM latent) distributions. The
model operates on $32\times32$ latent grids using patch size 1, with 
16 transformer layers and 16 attention heads. This high-capacity configuration 
allows the model to capture complex structural dependencies.
Unlike regression-based methods that minimize pixel-wise error, 
this generative formulation enables the model to produce realistic canopy
structures that preserve natural height distributions. Recent work
\cite{esser2024scaling} has demonstrated that rectified flow scales
predictably for high-resolution image synthesis, motivating our adoption
of this framework.

VibrantSR was trained in a staged manner. The Sentinel-2 encoder and CHM
autoencoder were each pretrained independently prior to training the flow
matching model, and their parameters were frozen in all subsequent
experiments.

The Sentinel-2 autoencoder was pretrained on Sentinel-2 imagery drawn
from the same geographic regions as the
downstream task. The overall architecture supports Sentinel-2 inputs at arbitrary temporal resolution or aggregations, 
including seasonal composites or single-date acquisitions, once trained for
the corresponding input distribution. Its purpose is to compress multi-spectral inputs into a latent representation that preserves spatial and spectral context relevant for canopy height inference, while reducing the dimensionality of the generative modeling problem.

The CHM autoencoder was pretrained separately on lidar-derived canopy
height tiles from the same study area. The decoder learns to reconstruct
high-resolution canopy height fields from latent representations that
capture fine-scale vertical structure. 

By pretraining both components
independently and freezing them thereafter, the flow matching network is
trained to learn a transport map between two fixed and compatible latent
manifolds. This staged training strategy isolates the contribution of the flow
matching model and avoids entangling representation learning with
generative transport, enabling more stable training and clearer
attribution of performance gains.

At inference time, the conditional Sentinel-2 latent is concatenated with an
input-conditioned noise vector, ensuring that identical inputs yield identical
outputs while retaining the expressivity of the generative model. $\frac{dz}{dt}=v_\theta(z,t)$ is then integrated from $z(0)=x$ to $t=1$ using the
\texttt{dopri5} ODE solver with 100 integration steps
\cite{lipman2024fmguide,chen2018neuralode}. No additional stochasticity is
introduced during integration, and outputs are fully reproducible for a given
input.

\subsection{Experimental Setup}\label{sec:experimental-setup}

We trained the flow matching model using AdamW with a batch size of 16.
Random horizontal and vertical flips, along with 90-degree rotations,
were applied on-the-fly to paired Sentinel-2 and CHM tiles before
encoding. The Sentinel-2 and CHM autoencoders were pretrained separately
and frozen during flow matching training. Training was performed on a
single node equipped with 8$\times$NVIDIA A100 GPUs. Total training compute
was approximately 5,400 A100-hours across all model components.

We compare \textbf{VibrantSR} against four baseline CHM products with differing input modalities, coverage, and spatial resolution (Table~\ref{tab:baselines}).

\begin{table}[htbp]
\centering
\caption{Baseline model comparison.}
\label{tab:baselines}
\begin{tabular}{llll}
\toprule
Model & Resolution & Input Data & Coverage \\
\midrule
VibrantVS \cite{chang2025vibrantvs} & 0.5~m & NAIP aerial & Western United States \\
Meta \cite{tolan2024very} & 1~m & Maxar Vivid2 & Global \\
LANDFIRE \cite{landfire2022} & 30~m & Landsat & United States \\
ETH \cite{lang2023high} & 10~m & Sentinel-2 & Global \\
\bottomrule
\end{tabular}
\end{table}

We evaluated performance using four metrics: Mean Absolute Error (MAE),
Mean Error (ME) indicating systematic bias, Block-R$^2$ measuring
explained variance at an aggregated spatial scale, and Edge Error (EE)
measuring structural fidelity via Sobel edge comparison \cite{tolan2024very}. Formal definitions are provided in the Appendix. Block-R$^2$ was
computed by aggregating predictions and ground truth over
non-overlapping $30\times30$ meter blocks in the CHM domain,
corresponding to approximately $15\,\mathrm{m}\times15\,\mathrm{m}$
ground area at 0.5~m resolution.
\section{Results}\label{sec:results}

\subsection{Quantitative Comparison}\label{sec:results-quant}

VibrantSR achieves an average Mean Absolute Error of 4.39 meters for canopy heights $\ge$2m across the test fold, outperforming
all satellite imagery-based models (Meta, LANDFIRE, and ETH; Table~\ref{tab:results}).
VibrantSR reduces MAE by 9\% relative to Meta (4.39~m vs 4.83~m), 26\%
relative to LANDFIRE (4.39~m vs 5.96~m), and 38\% relative to ETH (4.39~m
vs 7.05~m). VibrantSR also achieves substantially lower edge error than
these three models, indicating superior structural fidelity.

\begin{figure}[htbp]
\centering
\includegraphics[width=\textwidth]{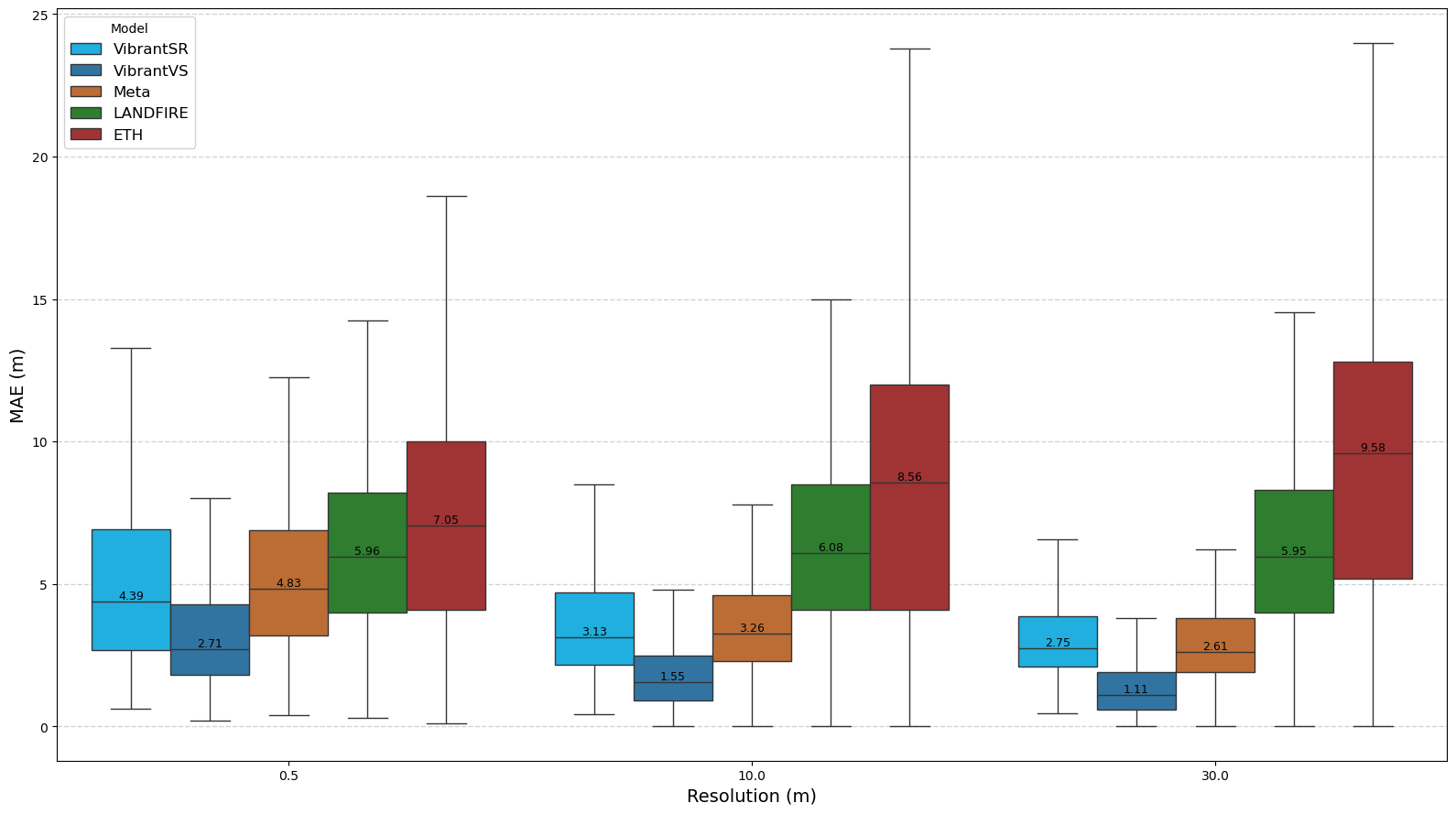}
\caption{Mean absolute error (MAE) as a function of output resolution for VibrantSR and baseline CHM products. Box–whisker plots summarize tile-level error at 0.5\,m, 10\,m, and 30\,m. VibrantSSR maintains stable accuracy across resolutions.}
\label{fig:mae}
\end{figure}

\begin{table}[htbp]
\centering
\caption{Performance comparison across models at 0.5~m resolution. All metrics computed for heights $\ge$2~m.}
\label{tab:results}
\begin{tabular}{lrrrr}
\toprule
Model & MAE & Mean Error & Block-R$^2$ & Edge Error \\
\midrule
\textbf{VibrantSR} & \textbf{4.39} & \textbf{-2.35} & \textbf{-0.62} & \textbf{0.16} \\
VibrantVS & 2.71 & -1.11 & 0.69 & 0.08 \\
Meta & 4.83 & -4.03 & -0.60 & 0.30 \\
LANDFIRE & 5.96 & 0.92 & -1.45 & 0.63 \\
ETH & 7.05 & 5.65 & -1.85 & 0.64 \\
\bottomrule
\end{tabular}
\end{table}

\begin{figure}[htbp]
\centering
\includegraphics[width=\textwidth]{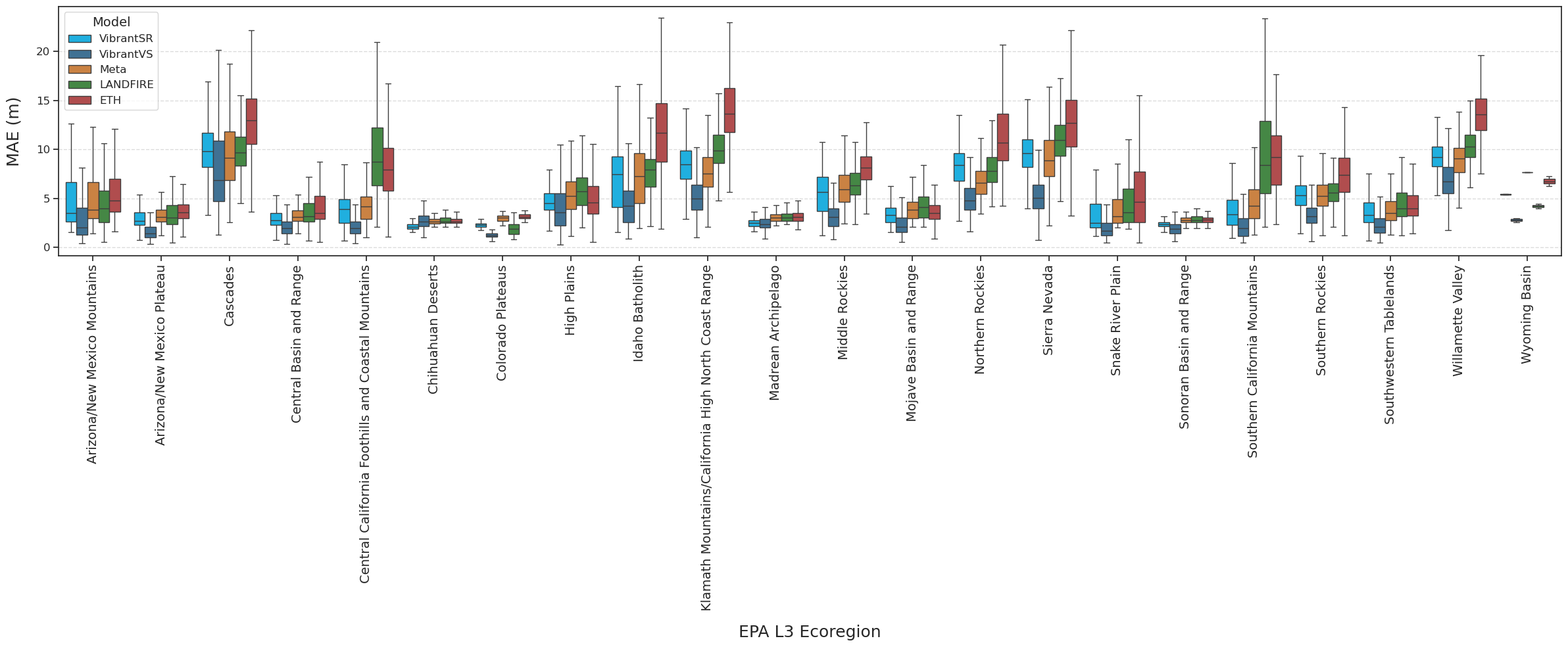}\\[0.6em]
\includegraphics[width=\textwidth]{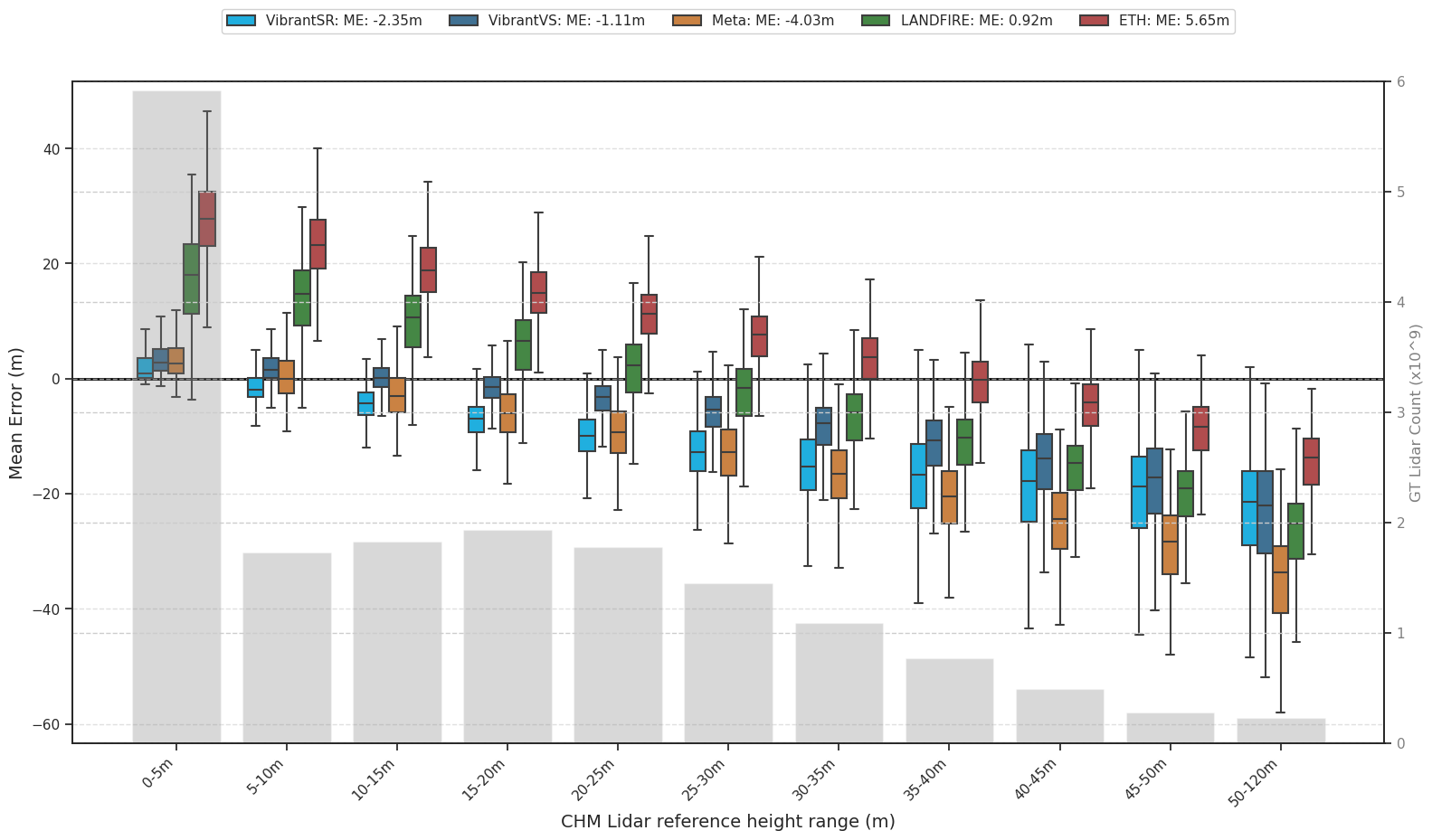}
\caption{(\textbf{Top}) Mean absolute error (MAE) by EPA Level-3 ecoregion for VibrantSR and competing CHM products. Each box–whisker summarizes the distribution of tile-level MAE within an ecoregion at 0.5~m resolution. VibrantSR is consistently among the lowest-error methods in structurally complex, high-biomass regions (e.g., Rockies, Sierra Nevada, Cascades), while performance converges in flatter and sparsely vegetated regions. 
(\textbf{Bottom}) Height-binned distribution of per-pixel canopy-height error ($\hat{h}-h$) at 0.5~m resolution. Boxes show the interquartile range, horizontal lines indicate the median, and whiskers denote the 5th--95th percentiles; grey bars give the relative lidar sample count. VibrantSR shows a smooth shift toward increasing underestimation with height, whereas LANDFIRE and ETH exhibit larger and less stable biases, particularly at low and very high canopy heights.}
\label{fig:ecoregion-and-bins}
\end{figure}

Performance improvement relative to other satellite imagery-based models is mostly consistent across diverse EPA Level 3 eco-regions (Fig.~\ref{fig:ecoregion-and-bins}). 
In addition to aggregate error metrics, we show residuals stratified by canopy height to assess how
errors vary across the height distribution (Fig.~\ref{fig:ecoregion-and-bins}).

\subsection{Qualitative Results}\label{sec:results-qual}

\begin{figure}[htbp]
\centering
\includegraphics[width=\textwidth,height=0.9\textheight,keepaspectratio]{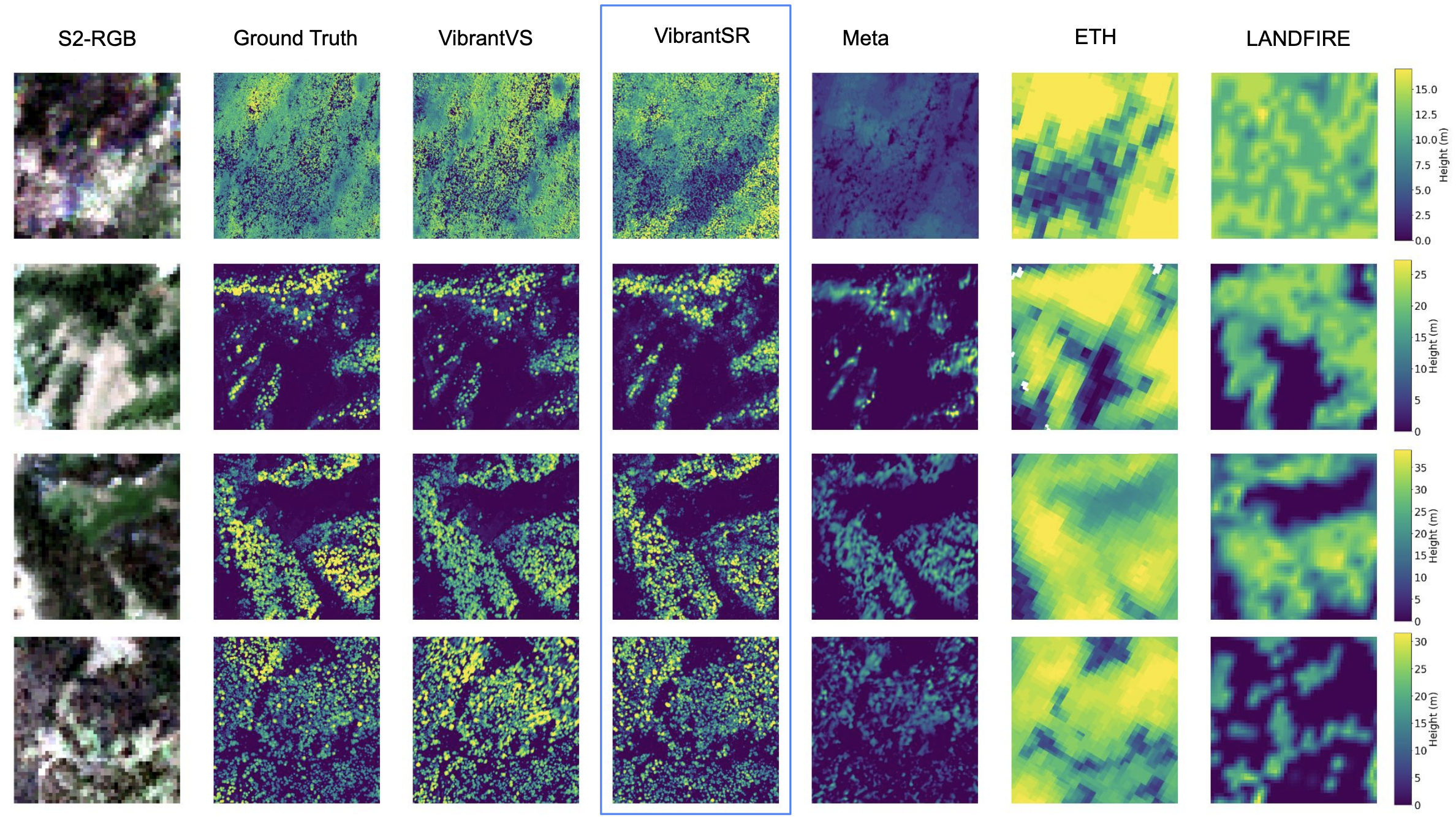}
\caption{Qualitative comparison of canopy-height predictions across multiple regions (one region per row). Within each row, all panels correspond to the same region; Sentinel-2 RGB is included among the \textbf{VibrantSR} input bands, whereas the other methods use their respective native input modalities.}
\label{fig:qual}
\end{figure}

\begin{figure}[htbp]
\centering
\begin{minipage}{0.4\textwidth}
    \centering
    \includegraphics[width=\textwidth]{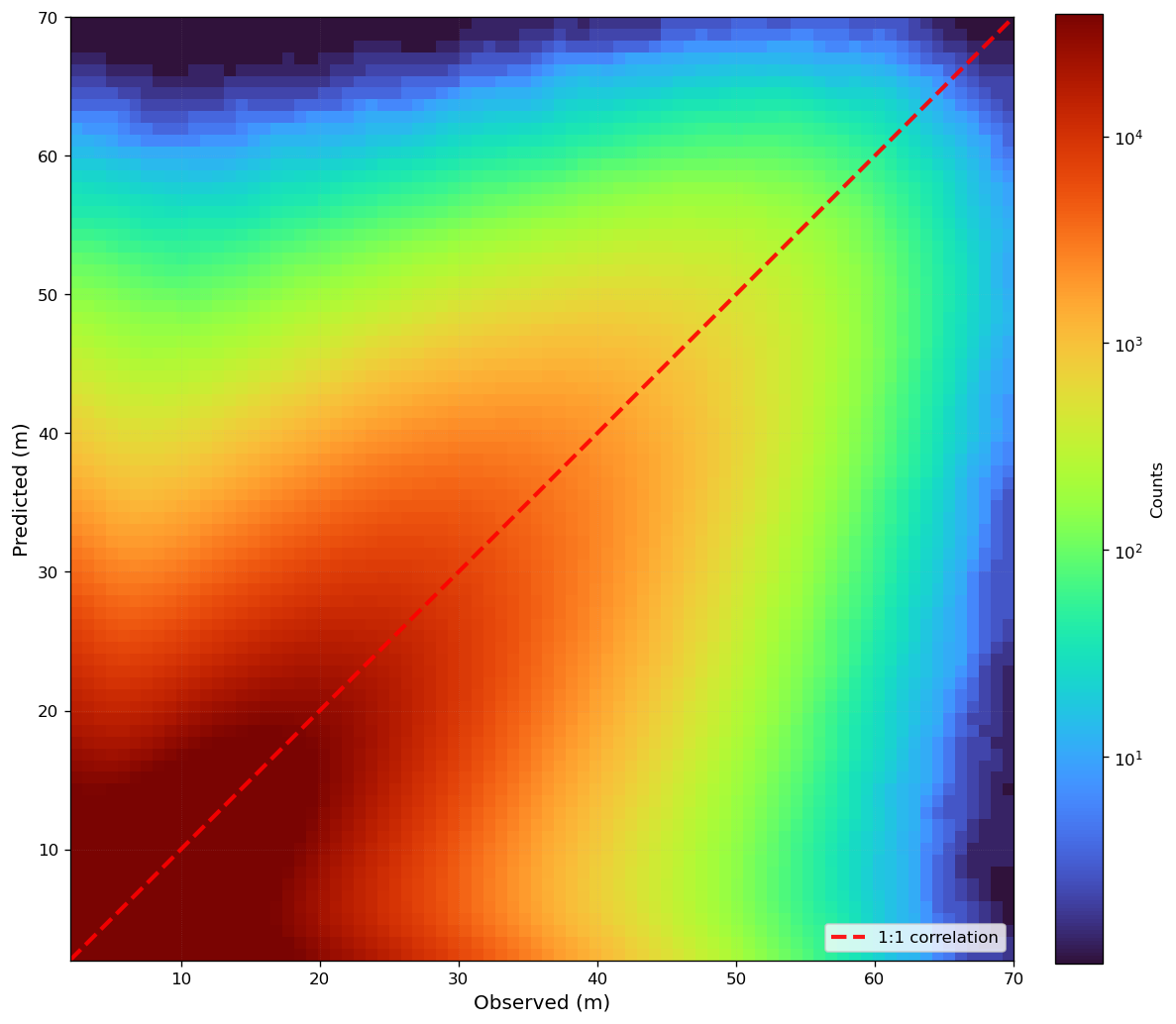}
\end{minipage}\hfill
\begin{minipage}{0.6\textwidth}
    \centering
    \includegraphics[width=\textwidth]{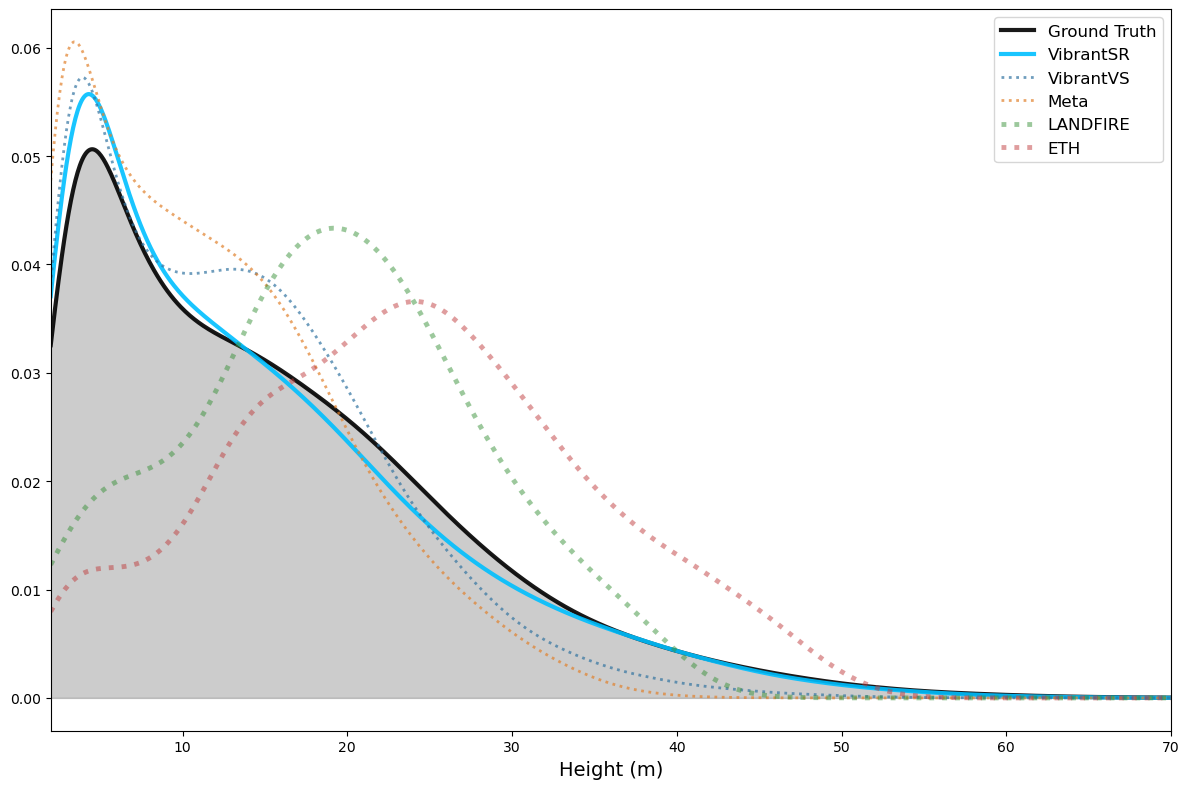}
\end{minipage}
\caption{Distributional agreement between lidar reference canopy heights and model predictions for pixels with reference CHM $\ge$2~m. (\textbf{Left}) Joint distribution of ground truth versus \textbf{VibrantSR} predictions. (\textbf{Right}) Kernel density estimates of canopy-height distributions for ground truth and predictions from \textbf{VibrantSR}, \textbf{VibrantVS}, Meta, LANDFIRE, and ETH.}
\label{fig:dist-agreement}
\end{figure}

Figure~\ref{fig:qual} shows side-by-side comparison of input
Sentinel-2 imagery (10~m), ground-truth lidar CHM (0.5~m), and
VibrantSR prediction (0.5~m) along with predictions from other models. VibrantSR uses coarse spectral inputs to
generate plausible sub-meter canopy structure---including gaps and
edges---that is typically smoothed by regression-based approaches.
VibrantSR also closely matches the ground-truth distribution of canopy
heights (Figs.~\ref{fig:dist-agreement}), 
indicating the generative approach captures
realistic structural variability rather than simply smoothing
predictions toward the mean.

\section{Discussion}\label{sec:discussion}

VibrantSR decouples high-resolution CHM estimation from aerial imagery,
enabling broad deployment with seasonal-to-annual update frequency. The
generative approach helps synthesize fine-grained canopy gaps and height
distributions that are often smoothed by regression-based methods, and
the model generalizes well across diverse forest types and climate
zones. This represents a significant advance over existing
satellite-based products: while ETH \cite{lang2023high} and LANDFIRE
\cite{landfire2022} provide broader coverage, their coarser resolution
(10--30~m) limits utility for fine-scale forest management decisions.
VibrantSR bridges this gap by achieving sub-meter output resolution
from freely available global imagery. However, VibrantVS (2.71~m MAE)
retains a significant accuracy advantage over VibrantSR (4.39~m MAE),
reflecting the fundamental information loss in 10~m vs 0.5 m input
resolution. This gap is unlikely to be closed through architectural
improvements alone; the spectral information in Sentinel-2 imagery quarterly aggregates
cannot uniquely determine sub-meter canopy structure. However, for many landscape-based uses,
the current residual remains acceptable.

The choice of generative flow matching over regression-based approaches
merits discussion. Traditional CHM models trained with MSE loss tend to
produce smoothed predictions that regress toward the mean losing the
high-frequency structural detail critical for applications such as
individual tree detection and canopy gap mapping. Our distribution
matching results (Figs.~\ref{fig:dist-agreement}) demonstrate
that VibrantSR preserves the natural variability of canopy heights, a
property that emerges from the generative formulation rather than
explicit structural losses. This suggests that flow matching may be
broadly applicable to remote sensing tasks where preserving
distributional properties is important. The negative Block-R$^2$
(Table~\ref{tab:results}) indicates that, under a pixel-wise
variance-explained metric, predictions do not match ground truth at the
chosen scale of spatial aggregation. We attribute this in part to a
double-penalty effect: when high-frequency structure is slightly shifted
relative to the label, pixel-wise metrics penalize it both as a false
positive and a false negative, which can drive $R^2$ negative even
when the predicted height distribution and texture are visually
plausible. For most of the intended use cases for VibrantSR, distributional agreement
(Figs.~\ref{fig:dist-agreement}) and
residual analysis are therefore
complementary and often more relevant indicators than strict pixel-level
correspondence.

\begin{figure}[htbp]
\centering
\includegraphics[width=\textwidth]{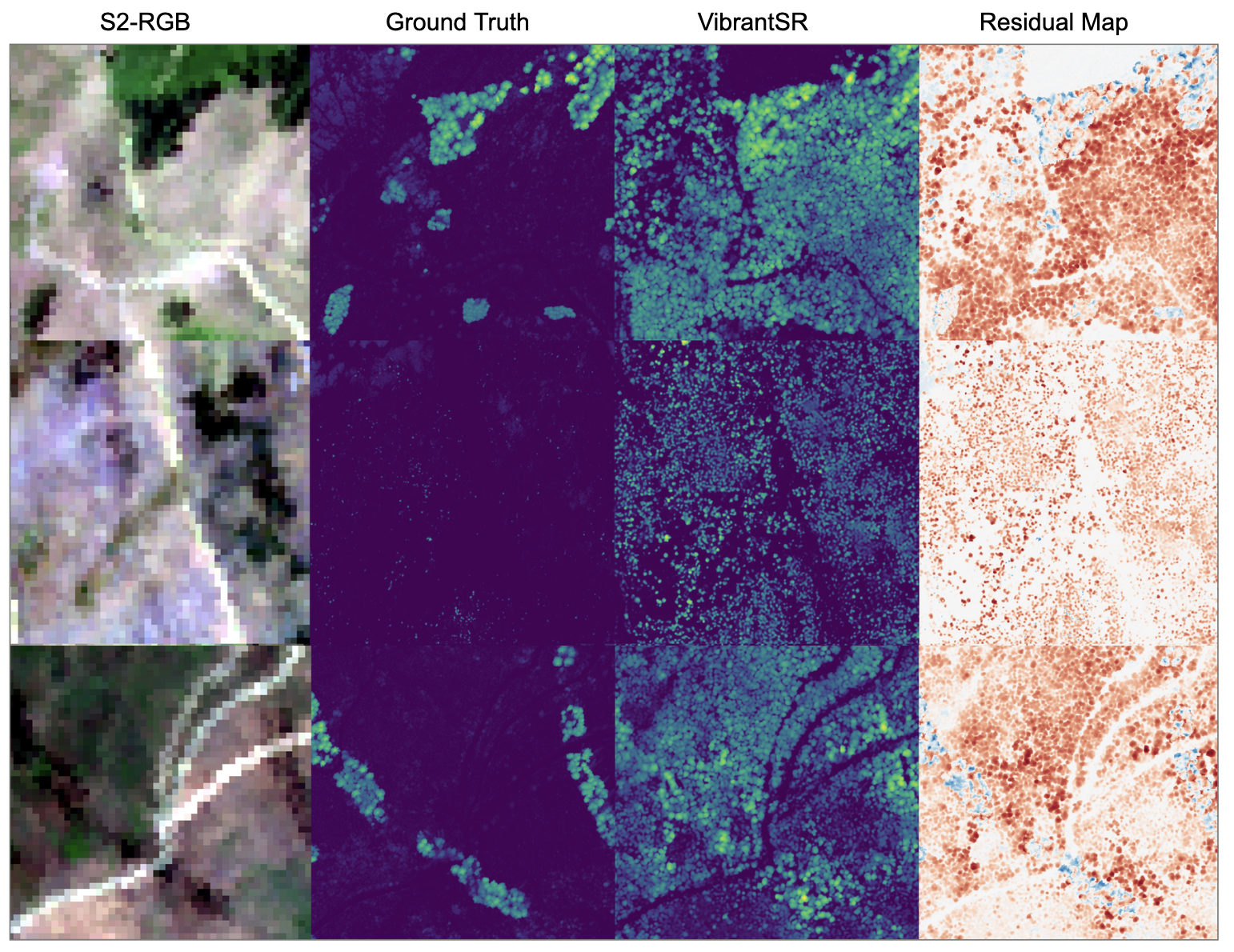}
\caption{Failure modes in regions containing tall trees. From left to right: 
color corrected 
RGB Sentinel-2 summer aggregate inputs, the ground-truth lidar-derived CHM, super-resolved synthetic 
CHM, and an error heatmap (reds=overestimation, blues=underestimation) displaying hallucinated height predictions.}
\label{fig:failure-modes}
\end{figure}

Several limitations should be noted when interpreting and applying
VibrantSR outputs. First, the evaluation is currently limited to the
western United States. While the eco-regions cover a diverse range of
forest types, performance in tropical or boreal forests with different
structural characteristics remains to be validated. Second, the temporal
mismatch between available lidar training data and the 2024 Sentinel-2
imagery introduces noise in both training and evaluation. Although we
prioritize recent lidar collections, the rapidly changing nature of
forest structure due to disturbance means that some training labels may
not accurately reflect the current state of the canopy seen by the
satellite. Third, the reliance on optical Sentinel-2 imagery means that
the model cannot ``see'' through clouds or dense smoke. While our
seasonal aggregation strategy mitigates this, persistent cloud cover in
some regions may result in data gaps or lower-quality predictions.
Qualitative inspection also reveals failure modes common to optical satellite-based monitoring (Fig.~\ref{fig:failure-modes}): model hallucination is observed primarily in areas with very tall, dense canopy where the Sentinel-2 signal saturates, and sharp boundaries at forest edges can appear blurred due to mixed-pixel effects. 

To reduce these effects in future work, we will (i) avoid median-composited inputs in favor of per-date Sentinel-2 observations to preserve fine spatial structure, (ii) leverage multiple temporally independent observations of the same location and regularize predictions to be consistent across acquisitions, and (iii) include additional supervision focused on regions and tiles where hallucinations are repeatedly observed.

A related constraint is that we intentionally avoid conditioning on GEDI and on other ancillary sensors in the current model. While GEDI provides valuable vertical-structure information, its sampling is sparse and uneven, and incorporating it would require a more complex multi-modal framework with careful spatiotemporal co-registration, explicit handling of missing data, and robustness to differing coverage patterns across regions and years. In practice, this can introduce additional sources of bias and complexity, and reduce deployability by coupling predictions to the availability and stability of an extra modality.

Finally, the 0.5~m output resolution, while structurally plausible, is a generative super-resolution from 10~m inputs. The fine-scale details are synthesized based on learned distributions and should be interpreted as statistically representative realizations rather than exact physical measurements. Outputs should not be treated as a substitute for field surveys or lidar in regulatory, legal, or safety-critical contexts, nor used for fine-scale operational decisions such as tree-level hazard assessment.

VibrantSR occupies a distinct niche in the landscape of canopy height
products, and practitioners should select among available options based on their specific requirements. For project-level planning requiring the highest accuracy---such as individual fuel break design or precise thinning prescriptions---VibrantVS or direct lidar acquisition remain preferable. However, for regional screening, prioritization, and monitoring applications where broad coverage and temporal consistency matter more than sub-meter precision, VibrantSR offers a practical solution. The seasonal-to-annual update capability enables before-and-after disturbance comparisons and multi-year trend analysis that are not possible with static products like Meta or LANDFIRE. The approach is well-suited as a screening layer to identify priority areas for targeted lidar acquisition or field validation, rather than as a standalone decision tool. By enabling sub-meter canopy height inference from freely available Sentinel-2 imagery, VibrantSR lowers barriers to accessing structurally detailed forest information, with particular relevance for regions with limited resources for airborne data acquisition.

\section{Conclusion}\label{sec:conclusion}
We introduced \textbf{VibrantSR}, a generative flow-matching approach that produces 0.5~m canopy height models (CHMs) from 10~m Sentinel-2 imagery. Across 22 EPA Level~3 ecoregions in the western United States, VibrantSR attains a mean absolute error of 4.39~m for reference canopy heights $\ge$2~m, corresponding to a 9--38\% reduction in error relative to widely used satellite-based CHM baselines (Meta, LANDFIRE, ETH). Beyond aggregate accuracy, the generative formulation better preserves canopy-height distributions and fine-scale structural detail, yielding lower edge error and closer agreement with lidar-derived reference statistics than regression-based alternatives.

VibrantSR helps address a key limitation in large-area forest monitoring: the scarcity of high-resolution structural estimates with consistent, repeatable temporal coverage. By avoiding dependence on aerial imagery acquisition, the method enables seasonal-to-annual CHM updates from globally available satellite observations. While aerial-imagery approaches such as VibrantVS remain preferable for site-specific analyses where airborne inputs are available, VibrantSR offers a practical capability for regional screening, change detection, and carbon accounting at continental scales.

Future work will broaden evaluation to additional forest biomes, train with multi-seasonal aggregates to support year-round inference, and incorporate auxiliary modalities (e.g., SAR) to further reduce error. We will also mitigate the dominant optical failure modes in tall, dense canopies and along sharp forest boundaries by (i) replacing median-composited inputs with per-date Sentinel-2 observations to better preserve fine spatial structure, (ii) leveraging multiple temporally independent acquisitions with consistency regularization across observations, and (iii) adding targeted supervision for regions and tiles where hallucinations recur.

\section{Author Contributions}\label{sec:contributions}
Conceptualization, K.N., A.G. and T.C.; methodology, K.N.; data curation, V.A.L. and L.J.Z.; writing—review and editing, K.N, A.G., D.D., N.E.R., T.C., V.A.L., and L.J.Z.; project administration, T.C., N.E.R., G.B and S.C. Author order for D.D., V.A.L, N.E.R. and L.J.Z. is alphabetical by last name.

\newpage
\section*{Data Availability}
\addcontentsline{toc}{section}{Data Availability}

Ground-truth canopy height models are derived from the dataset
introduced in VibrantVS \cite{chang2025vibrantvs}, which is publicly
available. Sentinel-2 Level-2A imagery is freely accessible via the
Copernicus Open Access Hub and Google Earth Engine. Model architecture, and
inference code will be made available upon publication.

\appendix
\section{Metric Definitions}\label{sec:appendix-metrics}

\textbf{Mean Absolute Error (MAE):}
\[\mathrm{MAE} = \frac{1}{n} \sum_{i=1}^{n} \left|y_i - \hat{y}_i\right|\]

\textbf{Mean Error (ME):}
\[\mathrm{ME} = \frac{1}{n} \sum_{i=1}^{n} \left(\hat{y}_i - y_i\right)\]

\textbf{Block-R$^2$:} 
\[R^2_{\text{block}} = 1 -
\frac{\sum_{b=1}^{B}\sum_{i \in b} \left(y_i - \hat{y}_i\right)^2}
{\sum_{b=1}^{B}\sum_{i \in b} \left(y_i - \bar{y}_b\right)^2}\]

\textbf{Edge Error (EE):}
\[\mathrm{EE} = \frac{1}{n} \sum_{i=1}^{n} \left|E(\hat{y}_i) - E(y_i)\right|\]
where $E(\cdot)$ represents the Sobel edge detection operation.

\newpage
\bibliography{references}

\end{document}